\documentclass{article}
\pdfoutput=1
\usepackage{spconf,amsmath,graphicx}
\usepackage{multirow}

\usepackage{microtype}
\usepackage{todonotes}

\title{Cross-Lingual Transfer Learning for Spoken Language Understanding}
\name{Quynh Ngoc Thi Do,  Judith Gaspers}
\address{Amazon\\Aachen, Germany}
\begin{document}
%
\maketitle
\vspace{-0.5cm}
\begin{abstract}
Typically, spoken language understanding (SLU) models are trained on annotated data which are costly to gather. Aiming to reduce data needs for bootstrapping a SLU system for a new language, we present  a simple but effective weight transfer approach using data from another language. The approach is evaluated with our promising multi-task SLU framework developed towards different languages. We evaluate our approach on the ATIS and a real-world SLU dataset, showing that i) our monolingual models outperform the state-of-the-art, ii) we can reduce data amounts needed for bootstrapping a SLU system for a new language greatly, and iii) while multi-task training improves over separate training, different weight transfer settings may work best for different SLU modules. 
\end{abstract}
\begin{keywords}
Spoken Language Understanding, Transfer Learning
\end{keywords}
\vspace{-0.2cm}
\section{Introduction}
Playing a crucial role in spoken dialogue systems, spoken language understanding (SLU) typically involves two sub-tasks of intent classification and slot filling. While the former identifies a speaker's intent, the latter extracts semantic constituents from the natural language query. Consider an example from the ATIS data \cite{conf/slt/TurHH10}: \textbf{city}~~~~~~ [$_{\bf{O}}$~where] [$_{\bf{O}}$~is] [$_{\bf{B-airport\_code}}$~MCO].
The slot filling sub-task should classify ``where'' and ``is'' as ${\bf{O}}$ and MCO as ${\bf{B-airport\_code}}$, the airport code. Meanwhile, the intent classification sub-task should identify \textbf{city} as the speaker's intent. 

Over the past few years, we have observed the success of deep neural networks (DNN) in SLU (e.g. \cite{conf/slt/TurHH10,Liu2016AttentionBasedRN,Zhang:2016:JMI:3060832.3061040, 7078634}). 
While traditionally separate models for intent detection and slot filling have been explored, recently there has been a shift towards joint models (e.g. \cite{Liu2016AttentionBasedRN, Zhang:2016:JMI:3060832.3061040, 7078634}) to leverage the interaction between the two tasks, which has been shown to improve performance.  
Typically, DNN models for SLU are trained on (large amounts of) annotated training data.

Due to the growing interest in devices making use of SLU technology, such as Amazon Alexa or Google Home, an important goal is porting SLU models to new languages in a quick and cost-efficient manner, i.e. without collecting large amounts of annotated training data. Towards this goal, in this paper we first present a flexible and modular multi-task SLU framework that supports various deep learning architectures, including recent techniques which have shown promising results on related tasks already. The framework provides an easy way to select the best monolingual model for a given target language and training data size, as the behavior of different deep learning technique might differ accordingly \cite{C18-2035}.

We then explore leveraging data from another language, assuming that a SLU system for this language is already available. In particular, we train a DNN on data from one language and use its weights to initialize a DNN for another (target) language -- an approach also known as \textit{transfer learning}. While transfer learning has already been shown to be effective in several tasks, including slot filling/named entity recognition \cite{yang2017,riedl2018}, to the best of our knowledge it has not yet been explored for joint intent detection and slot filling. Previous work on porting SLU models has mainly explored using MT (e.g. \cite{gaspers2018,garcia2012,he2013}). Recently, \cite{zero} presented a zero-shot and a bilingual training approach, where the latter requires an intensive modification to an existing SLU system, contrasting with our simple weight transfer approach which can be easily applied to any existing neural network based SLU system. 

We evaluate our approach both on the ATIS benchmark dataset and on a real-world SLU dataset. The main contributions of this paper are: i) we propose a flexible multi-task SLU system, which outperforms the state of the art on ATIS, ii) we explore different approaches for transferring weights, and iii) we show that our approach allows large data reductions on a real-world dataset while keeping performance. 
\vspace{-0.2cm}
\section{Multi-task SLU Framework}
\begin{figure}[htb]
  \centering
  \includegraphics[width=8cm]{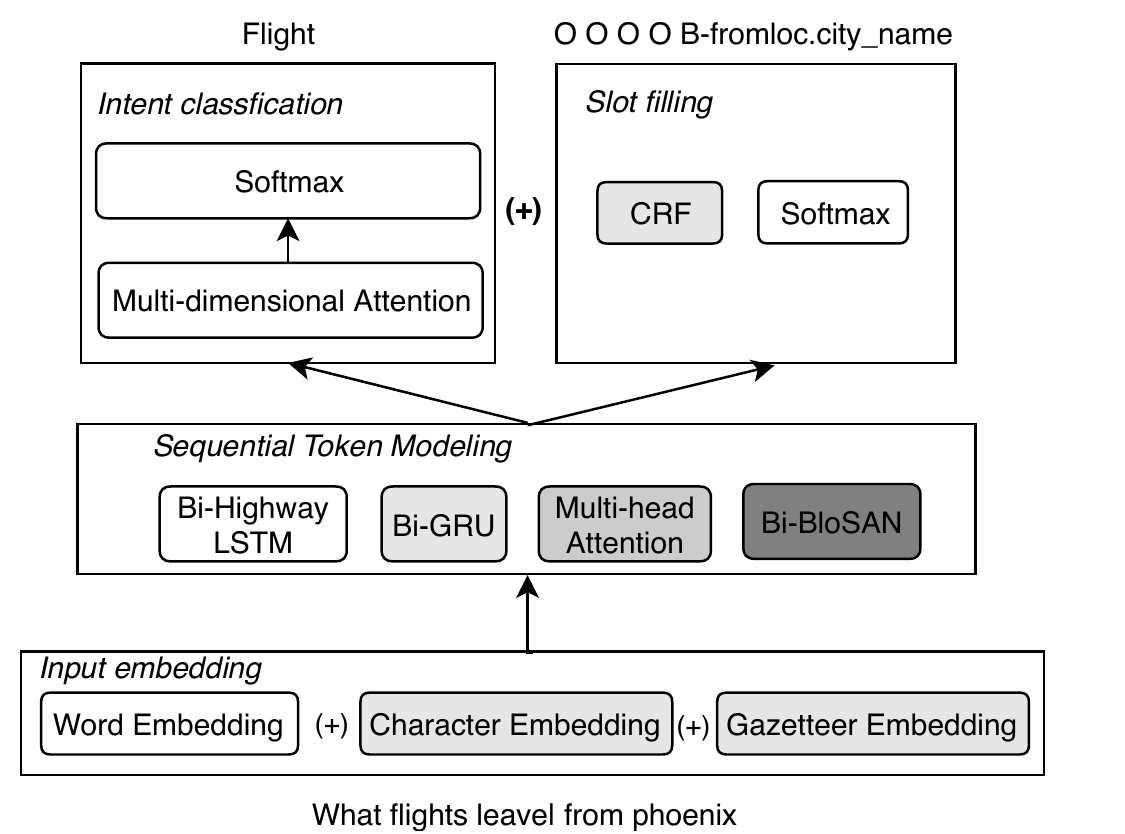}
  \caption{System Architecture}
\label{fig:architecture}
\vspace{-0.2cm}
\end{figure}
As shown in Fig. \ref{fig:architecture}, the framework divides model construction into four phases:
(I) input embedding, (II) sequential token modelling, (III) slot filling, and (IV) intent classification.
\vspace{-0.3cm}
\paragraph*{Input Embedding} The input embedding of a token consists of three optional concatenated components: a word embedding, a character embedding, and a gazetteer embedding. 
\textit{First}, the word embedding layer is either initialized by random vectors or pre-trained word embeddings. \textit{Second}, the character embedding is learned by a 1-dimensional convolution neural network (CNN) on the tokens' character sequences. 
\textit{Third}, given a list of gazetteer types, where each type $g_i$ contains a list of gazetteer names ${p_i}_j$.
Each token in an utterance is assigned an integer number as gazetteer feature.  
If a phrase ${p_i}_j \in g_i$ matches a sub-string of the utterance, the first word and the remaining words of the matched string will receive $2*i-1$ and  $2*i$ as the gazetteer features, respectively. If a token does not occur in any matched sub-string, it will have $0$ as gazetteer feature.\footnote{Note that if there are more than one match at a particular token, only the longest match is considered.}
After the concatenation, this phase produces a fixed-dimensional real vector for each token.
\vspace{-0.3cm}
\paragraph*{Sequential Token Modeling} As in the first phase, this phase also produces a fixed-dimensional representation for each token, but by taking into account the contextual information from other tokens in the utterance.
We propose three different architectures for this phase: Bi-directional RNN, Attention and Bi-directional Attention.
\begin{figure}
\centering
\begin{minipage}{.25\textwidth}
  \centering
  \includegraphics[width=0.8\linewidth]{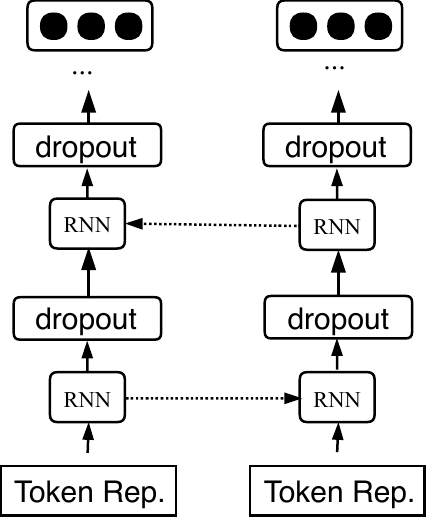}
  \caption{RNN Architecture.}
  \label{fig:rnn}
\end{minipage}%
\begin{minipage}{.25\textwidth}
  \centering
  \includegraphics[width=0.65\linewidth]{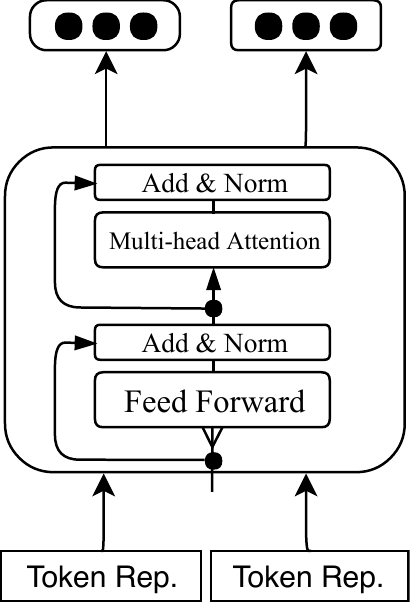}
  \caption{Attention Architecture.}
  \label{fig:att}
\end{minipage}
\vspace{-0.3cm}
\end{figure}
\textit{First}, Fig. \ref{fig:rnn} shows the \textit{bi-directional RNN architecture} of the sequential token modeling module. The RNN unit can be either a GRU or a highway LSTM \cite{NIPS2015_5850} with recurrent dropout \cite{Gal:2016:TGA:3157096.3157211} as proposed in \cite{P17-1044}. The output of the top-most RNN layer is the output of this module. 
\textit{Second}, the \textit{attention architecture} of the sequential token modeling module is shown in Fig. \ref{fig:att}, which is similar to the well-known multi-head attention applied in machine translation recently \cite{Vaswani2017AttentionIA}.
\begin{figure}[htb]
  \centering
  \includegraphics[width=5.5cm]{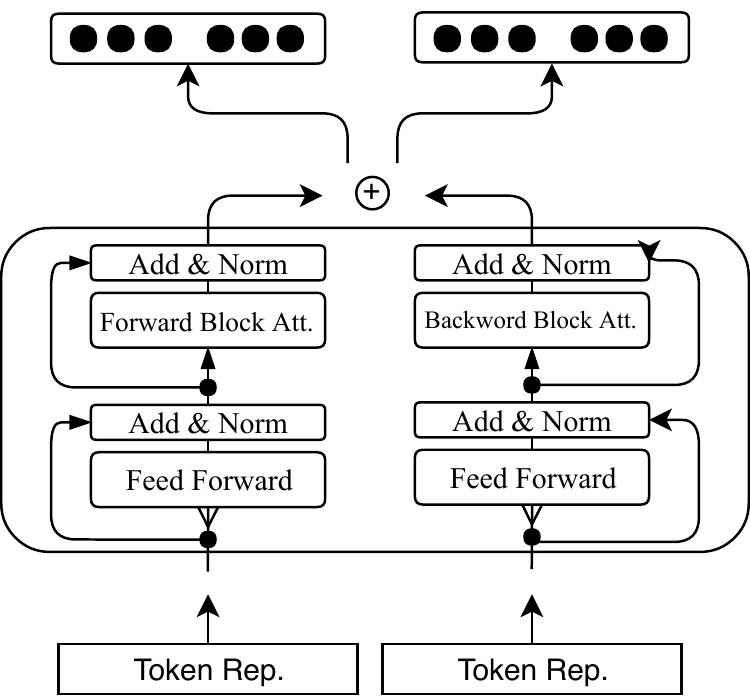}
  \caption{Bi-directional Attention Architecture.}
\label{fig:block-att}
\vspace{-0.3cm}
\end{figure}
\textit{Third}, in Fig. \ref{fig:block-att}, we propose an architecture to deal with a bi-directional attention mechanism which is the bi-block multidimensional attention \cite{shen2018bidirectional} in our implementation. It consists of two parallel sub-networks for forward and backward directions, in which each sub-network is similar to the attention network as seen in Fig. \ref{fig:att}.
\vspace{-0.3cm}
\paragraph*{Intent Classification}
\begin{figure}
\centering
\begin{minipage}{.25\textwidth}
  \centering
  \includegraphics[width=0.5\linewidth]{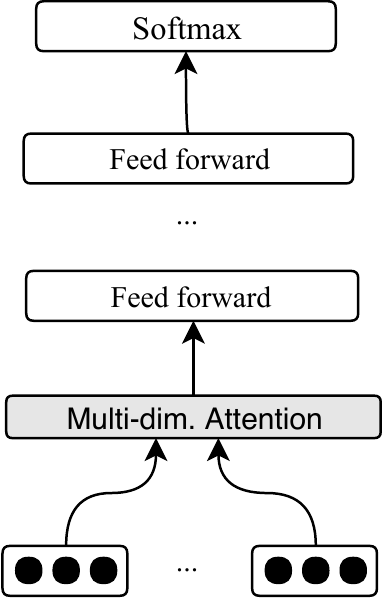}
  \caption{Intent classification.}
  \label{fig:intent}
\end{minipage}%
\begin{minipage}{.25\textwidth}
  \centering
  \includegraphics[width=0.85\linewidth]{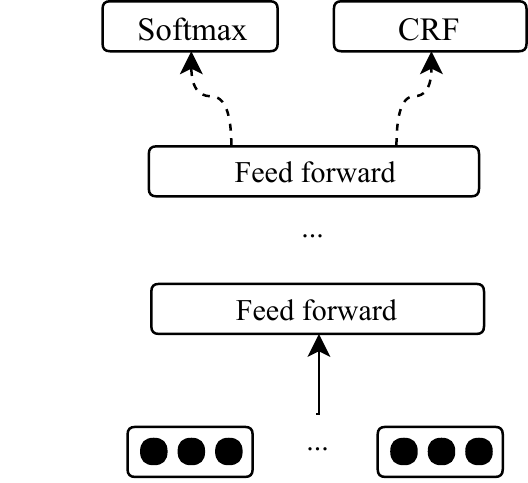}
  \caption{Slot filling.}
  \label{fig:slot}
\end{minipage}
\vspace{-0.3cm}
\end{figure}
The intent prediction's architecture is shown in Fig. \ref{fig:intent}. This phase consists of three main components: a multi-dimensional attention layer, a feed-forward network and a softmax layer. It first receives as inputs the contextual representations of the tokens from the previous phase. Then, a dimensional attention layer is applied on the tokens' representation to obtain a single representation for each utterance. Finally, the utterance representations are passed to a stack of a feed-forward network and a softmax layer to compute the intent distribution. To improve the performance, we apply label smoothing \cite{Szegedy2016RethinkingTI} in this phase.
\vspace{-0.3cm}
\paragraph*{Slot Filling}
Fig. \ref{fig:slot} shows the slot filling architecture. It is served by the sequential token representations as inputs. A feed-forward network with a softmax or a linear conditional random field (CRF) on top, is used to compute the slot distributions. Label smoothing \cite{Szegedy2016RethinkingTI} can be applied with the softmax in this phase.

\vspace{-0.3cm}
\paragraph*{Multi-Task System}
The intent prediction and slot  filling sub-tasks can be trained jointly or separately via the following combined loss function: $L = \alpha_i \hat{L}_i + \alpha_s \hat{L}_s$, 
where $\alpha_i, \alpha_s$ are the weights indicating the importance of the intent prediction and slot filling, respectively. $\hat{L}_i, \hat{L}_i$ are the normalized form of $L_i$ and $L_s$ respectively. 
\section{Experiments}
In the following, we will first describe our datasets and then compare our SLU framework on the ATIS dataset to the state-of-the-art. Subsequently, we explore transfer learning from English to German both on ATIS and on a real-world dataset.
\vspace{-0.3cm}
\subsection{Datasets}
The ATIS dataset \cite{conf/slt/TurHH10} has been widely used in SLU research. It contains audio recordings and corresponding annotated transcriptions in English of people making flight reservations. In our experiments, we use the version provided by \cite{N18-2118}, in which the training, development and test sets contain 4.478, 500 and 893  utterances, respectively. 
For language transferring experiments, we translated the test set, 463 random utterances from the training set, and 144 random utterances from the development set into German.

To evaluate our approach in a real-world scenario, we extracted a random sample of 1M training data utterances from a deployed large-scale English SLU system as well as random samples of 10k and 20k from a German system for training and 2k to create a development set. These utterances are representative of user requests to voice-controlled devices and cover a large number of different slots and intents.

We collect from our internal database the lists of city names, airport names, airline names and airline codes to be used as gazetteers.

For evaluation we use the standard metrics, i.e. F1, precision and recall for slot filling (computed using the CoNLL 2002 script) and accuracy for intent classification. 
\vspace{-0.3cm}
\subsection{Monolingual models on benchmark ATIS}
To compare our approach to the state-of-the-art, we first evaluate our models on ATIS data; in this experiment we use GloVe \cite{Pennington2014} 100-dimensional word embeddings. For character embeddings, characters are embedded in 8-dimensional embeddings. The convolutions have window sizes of 3, 4, and 5
characters, each consisting of 50 filters. The sequential token modeling has the depth of 2 in all architectures. The hidden layers in the RNN architecture are of size 300, while the number of heads in the multi-head attention and number of blocks in the bi-block multi-dimensional attention are set to 2 and 3, respectively. The feed-forward networks in slot filling and intent classification both consist of two layers of size 300. Dropout keep probability is set to 0.9 in all the cases except for the residual dropout in multi-head attention where it is set to 0.8. $\alpha_i, \alpha_s$ and the label smoothing rate are tuned on the development set resulting in $\alpha_i=0.2, \alpha_s=0.8$ and label smoothing rate $ =0.1$.   We train our models using Adam optimizer with 0.001 as the learning rate.   In line with previously reported results, we do not use external knowledge (gazetteers) and average the scores of 5 runs for each experiment. The results are presented in Table \ref{tbl:basic}.
\begin{table}[!ht]
\centering
\resizebox{0.48 \textwidth}{!} {
\begin{tabular}{l|lll|l}
\multirow{2}{*}{\textbf{Model}} & \multicolumn{3}{c}{\textbf{Slot}}     & \textbf{Intent} \\ 
                                & \textit{P} & \textit{R} & \textit{F1} & \multicolumn{1}{c}{\textit{Acc.}}   \\\hline \hline
Hakkani-Tur et al., 2016  \cite{multijoint}                       &            &            &       94.3      &          92.6       \\
Liu and Lane, 2016 \cite{Liu2016AttentionBasedRN}                          &            &            &   94.2          &  91.1\\            
Goo et al., 2018  \cite{N18-2118}                        &            &            &  95.2         &  94.1\\     \hline
Highway:W                          &      95.4      &  95.3            &   95.4      &  96.5 \\ 
Highway:CNN                          &      94.5      &  94.1            &   94.3      &  95.8 \\ 
Highway:W+CNN                          &      95.7      &  95.6            &   \textbf{95.6 }     &  96.8 \\ 

GRU:W+CNN                        &   95.2         &      95.3      &    95.2     & 96.8 \\ 

MulHeadAtt:W+CNN                        &  93.7          &    94.3        &     94.0    &  \textbf{97.0} \\ 
Block-Dim. Att:W+CNN                         &     93.9       &    94.6        &      94.3   &  96.8\\ 
\end{tabular}}
\caption{Different models in basic setting compared to the state-of-the-art results borrowed from \cite{N18-2118}. W--Word embeddings, CNN -- CNN character embeddings}
\label{tbl:basic}
\vspace{-0.2cm}
\end{table}
Overall, our models outperform the state-of-the-art. For intent detection gains are comparatively large, yielding up to 97.0 in accuracy, which implies a gain of 2.9 absolute compared to the previously best reported result of 94.1. For slot filling, improvements are lower, but several of our models still outperform the state-of-the-art, yielding up to 95.6 in F1 compared to the previously reported 95.2. Due to best performance on slot filling and competitive performance on intent detection, we use \textit{Highway:W+CNN} for more detailed analyses and subsequent experiments.

To explore whether our multi-task system achieves better results compared to training the intent detection and slot filling models separately, we ran separate training. With an F1 of 95.6 vs 95.4 and an intent accuracy of 96.8 vs 95.9 for joint vs separate training, respectively, in line with previously reported results, joint training improves results.

Recall that our frameworks supports applying either a CRF or softmax for slot filling. While a CRF is typically more accurate, softmax is quicker. 
Since experiments on NLP tasks imply that softmax can be similarly accurate as a CRF when it's applied together with label smoothing \cite{Szegedy2016RethinkingTI}, we investigated whether this also holds for our SLU framework. Results are presented in Table \ref{tbl:sm}.

\begin{table}[!ht]
\centering
\resizebox{0.48 \textwidth}{!} {
\begin{tabular}{l|lll|l}
\multirow{2}{*}{\textbf{Model}} & \multicolumn{3}{c}{\textbf{Slot}}     & \textbf{Intent} \\ 
                                & \textit{P} & \textit{R} & \textit{F1} & \multicolumn{1}{c}{\textit{Acc.}}   \\\hline \hline
CRF                       &      95.2   &  95.7         & 95.4     &  \textbf{96.8} \\ 

Softmax                       &      95.2   &  95.5           & 95.3     &  96.6 \\ 
Softmax + Lbl. Smoothing                        &     95.7      &  95.6            &   \textbf{95.6 }     &  \textbf{96.8} \\ 
\end{tabular}}
\caption{With vs. without label smoothing vs. CRF.}
\label{tbl:sm}
\vspace{-0.3cm}
\end{table}
The CRF outperforms single softmax, but not softmax with label smoothing. As a CRF is usually slower than softmax in both training and prediction, we propose to use softmax and label smoothing. Notice that speed is an important issue for large-scale industry SLU systems. 

To evaluate the effectiveness of using gazetteers, we train our full model \textit{Highway:W+CNN+G} using the internal gazetteers\footnote{Gazetteers embeddings are of size 50.} resulting in a slight improvement with 95.7 F1 for slot filling and 96.8 Acc. for intent classification. 
\vspace{-0.3cm}
\subsection{Cross-lingual transfer learning on ATIS}
To explore transfer learning on ATIS, we train our  \textit{Highway:W+CNN+G model}  using fixed MUSE multilingual embeddings \cite{conneau2017word} 
on the English ATIS data (except for the samples which 
are in parallel with our German ATIS) and select the best weights using the German development data. The weights are then used to initialize training on German ATIS data.  Since to the best of our knowledge transferring weights in a SLU multi-task system has not yet been explored, it is unclear which of the weights to transfer. Therefore, we explored different weight transferring settings on the German ATIS test set. The settings are listed in Table \ref{tbl:setting}.
\begin{table}[ht]
\begin{tabular}{ll}
Setting & Weights pre-trained by using English data   \\ \hline \hline
All   &  All phases\\
Full-Slot & All except slot filling\\
Full-Multidim & All except multi-dimensional att.\\
Full-bi-LSTM & All except sequential token modelling \\
\end{tabular}
\caption{Weight transfer settings for German ATIS model.}
\label{tbl:setting}
\vspace{-0.3cm}
\end{table}

To evaluate the gain from transferring weights, we created a monolingual baseline, i.e. we trained a model solely on the German ATIS data. To investigate whether performance is reasonable on German, we additionally trained a model on the parallel English data (i.e. the subset which was translated). With an F1 of 90.5 vs 89.6 and an intent accuracy of 88.4 vs 87.8 for English vs German, respectively, performance appears to be reasonable. Table  \ref{tbl:wtf_deatis1} presents how the weight transfer settings compare to the baseline model.
\begin{table}[ht]
\centering
\begin{tabular}{l|lll|l}
\multirow{2}{*}{\textbf{Model}} & \multicolumn{3}{c}{\textbf{Slot}}     & \textbf{Intent} \\ 
                                & \textit{P} & \textit{R} & \textit{F1} & \multicolumn{1}{c}{\textit{Acc.}}   \\\hline \hline
Monolingual &90.6 & 88.6 &89.6 &87.8\\
\hline
All &92.4 & 92.3 & \textbf{92.3} &89.0 \\ 
Full-Slot & 85.6 & 89.0 & 88.8 & 87.6 \\
Full-Multidim &90.4 & 90.2 & 90.3 & \textbf{89.5} \\
Full-bi-LSTM & 87.8 & 88.3 & 88.0 & 85.8 \\
\end{tabular}
\caption{Monolingual model vs. weight transfer results on German ATIS translations.}
\label{tbl:wtf_deatis1}
\vspace{-0.3cm}
\end{table}
The results show that transferring weights can improve performance, depending on which weights are transferred. Overall, taking gains in slot filling and intent detection together, the best setting is transferring all weights, which improves results for F1 from 89.6 to 92.3 and for intent detection from 87.8 to 89.5. 
However, for intent detection better results are achieved by \textit{full-multidim}, yielding an accuracy of 89.5. 
Notably though, in our framework the two tasks are sharing only token representations and can be easily separated even after joint training. 
Since we explore settings where only small data amounts in the target language are available, it can be  feasible to train one (potentially large-scale) source model and transfer it twice using different approaches, i.e. \textit{full} and \textit{full-multidim}, and then separate the modules to use the intent detection module transferred with \textit{full-multidim} and the slot filling module transferred with \textit{full}.    
\vspace{-0.3cm}
\subsection{Cross-lingual transfer learning on real-world data}
To explore potential data reductions in a real-world setting, we trained  baseline models on the 10k and 20k DE training datasets. In addition, we trained a model on the 1M EN utterances and transferred weights using the best-performing approach from the previous section, i.e. transferring all weights. As word embeddings, we used the fixed MUSE multilingual embeddings \cite{conneau2017word}. Results are presented in Table \ref{tbl:wtf_rw}.
\begin{table}[ht]
\centering
\resizebox{0.48 \textwidth}{!} {
\begin{tabular}{l|l|lll|l}
\multirow{2}{*}{\textbf{Data}} & \multirow{2}{*}{\textbf{Model}} & \multicolumn{3}{c}{\textbf{Slot}}     & \textbf{Intent} \\ 
                  &              & \textit{P} & \textit{R} & \textit{F1} & \multicolumn{1}{c}{\textit{Acc.}}   \\\hline \hline
10k DE& Monoling.       & 77.1 & 73.6 & 75.3 & 87.9  \\ 
10k DE, 1M EN& Transfer    &  79.2 &77.1 & \textbf{78.1} &\textbf{89.5} \\ 
\hline
20k DE    &   Monoling.      & 80.1 & 77.1 & 78.6 & 89.1 \\ 
20k DE, 1M EN& Transfer &82.6 & 80.5 & \textbf{81.5} & \textbf{90.4} \\
\end{tabular}}
\caption{Transfer learning results on a real-world dataset.}
\label{tbl:wtf_rw}
\end{table}
The results show gains for transferring weights on both datasets and for both intent detection and slot filling. For intent classification, training on 10k DE data with transferring weights outperforms training solely on 20k DE utterances (89.5 vs 89.1 in accuracy), despite using 50\% less DE data, indicating that by using cross-lingual transfer learning we can reduce data amounts needed for bootstrapping a large-scale SLU system greatly. While there are also gains for slot filling, training on 10k with transfer learning does not outperform training a model solely on 20k. 
More fine-grained analyses with different data sizes are needed to draw more precise conclusions on potential data reductions both for intent detection and slot filling, which we leave for future work.
\vspace{-0.3cm}
\section{Conclusion}
We presented a flexible and modular multi-task framework for intent detection and slot filling. With the framework, we compared different weight sharing settings for transferring knowledge from English to German. We presented results on the ATIS and a real-world dataset, showing that i) our models outperform the state-of-the-art, ii) we can reduce data amounts needed for bootstrapping a SLU system for a new language greatly by utilizing data from another language, and iii) while multi-task training improves over separate training, different weight transfer settings work best for intent detection and slot filling. Since our framework allows easy separation of modules even after multi-task training, it can be easily used to transfer with different settings and separate modules afterwards to get modules with best performance for application.


\bibliographystyle{IEEEbib}

\begin{thebibliography}{10}

\bibitem{conf/slt/TurHH10}
G{\"{o}}khan T{\"{u}}r, Dilek Hakkani-T{\"{u}}r, and Larry~P. Heck,
\newblock ``What is left to be understood in atis?,''
\newblock in {\em SLT}, Dilek Hakkani-T{\"{u}}r and Mari Ostendorf, Eds. 2010,
  pp. 19--24, IEEE.

\bibitem{Liu2016AttentionBasedRN}
Bing Liu and Ian Lane,
\newblock ``Attention-based recurrent neural network models for joint intent
  detection and slot filling,''
\newblock in {\em INTERSPEECH}, 2016.

\bibitem{Zhang:2016:JMI:3060832.3061040}
Xiaodong Zhang and Houfeng Wang,
\newblock ``A joint model of intent determination and slot filling for spoken
  language understanding,''
\newblock in {\em IJCAI'16}. 2016, pp. 2993--2999, AAAI Press.

\bibitem{7078634}
D.~Guo, G.~T{\"{u}}r, W.~Yih, and G.~Zweig,
\newblock ``Joint semantic utterance classification and slot filling with
  recursive neural networks,''
\newblock in {\em 2014 IEEE SLT Workshop}, Dec 2014, pp. 554--559.

\bibitem{C18-2035}
Quynh Ngoc~Thi Do, Artuur Leeuwenberg, Geert Heyman, and Marie-Francine Moens,
\newblock ``A flexible and easy-to-use semantic role labeling framework for
  different languages,''
\newblock in {\em COLING}. 2018, pp. 161--165, Association for Computational
  Linguistics.

\bibitem{yang2017}
Zhilin Yang, Ruslan Salakhutdinov, and William~W Cohen,
\newblock ``Transfer learning for sequence tagging with hierarchical recurrent
  networks,''
\newblock {\em arXiv preprint arXiv:1703.06345}, 2017.

\bibitem{riedl2018}
Martin Riedl and Sebastian Pad{\'o},
\newblock ``A named entity recognition shootout for german,''
\newblock in {\em ACL}, 2018, vol.~2, pp. 120--125.

\bibitem{gaspers2018}
Judith Gaspers, Penny Karanasou, and Rajen Chatterjee,
\newblock ``Selecting machine-translated data for quick bootstrapping of a
  natural language understanding system,''
\newblock {\em NAACL-HLT}, 2018.

\bibitem{garcia2012}
Fernando Garc{\'{\i}}a, Llu{\'{\i}}s~F. Hurtado, Encarna Segarra, Emilio
  Sanchis, and Giuseppe Riccardi,
\newblock ``Combining multiple translation systems for spoken language
  understanding portability,''
\newblock in {\em 2012 {IEEE} Spoken Language Technology Workshop (SLT), Miami,
  FL, USA, December 2-5, 2012}, 2012, pp. 194--198.

\bibitem{he2013}
X.~He, L.~Deng, D.~Hakkani-T{\"{u}}r, and G.~T{\"{u}}r,
\newblock ``Multi-style adaptive training for robust cross-lingual spoken
  language understanding,''
\newblock in {\em 2013 IEEE International Conference on Acoustics, Speech and
  Signal Processing}, May 2013, pp. 8342--8346.

\bibitem{zero}
Shyam Upadhyay, Manaal Faruqui, G{\"{o}}khan T{\"{u}}r, Dilek~Z.
  Hakkani{-}T{\"{u}}r, and Larry~P. Heck,
\newblock ``(almost) zero-shot cross-lingual spoken language understanding,''
\newblock in {\em 2018 {IEEE} International Conference on Acoustics, Speech and
  Signal Processing, {ICASSP} 2018, Calgary, AB, Canada, April 15-20, 2018},
  2018, pp. 6034--6038.

\bibitem{NIPS2015_5850}
Rupesh~K Srivastava, Klaus Greff, and J\"{u}rgen Schmidhuber,
\newblock ``Training very deep networks,''
\newblock in {\em Advances in Neural Information Processing Systems 28},
  C.~Cortes, N.~D. Lawrence, D.~D. Lee, M.~Sugiyama, and R.~Garnett, Eds., pp.
  2377--2385. Curran Associates, Inc., 2015.

\bibitem{Gal:2016:TGA:3157096.3157211}
Yarin Gal and Zoubin Ghahramani,
\newblock ``A theoretically grounded application of dropout in recurrent neural
  networks,''
\newblock in {\em NIPS}, USA, 2016, pp. 1027--1035, Curran Associates Inc.

\bibitem{P17-1044}
Luheng He, Kenton Lee, Mike Lewis, and Luke Zettlemoyer,
\newblock ``Deep semantic role labeling: What works and what's next,''
\newblock in {\em Proceedings of the 55th Annual Meeting of the Association for
  Computational Linguistics (Volume 1: Long Papers)}. 2017, pp. 473--483,
  Association for Computational Linguistics.

\bibitem{Vaswani2017AttentionIA}
Ashish Vaswani, Noam Shazeer, Niki Parmar, Jakob Uszkoreit, Llion Jones,
  Aidan~N. Gomez, Lukasz Kaiser, and Illia Polosukhin,
\newblock ``Attention is all you need,''
\newblock in {\em NIPS}, 2017.

\bibitem{shen2018bidirectional}
Tao Shen, Tianyi Zhou, Guodong Long, Jing Jiang, and Chengqi Zhang,
\newblock ``Bi-directional block self-attention for fast and memory-efficient
  sequence modeling,''
\newblock in {\em International Conference on Learning Representations}, 2018.

\bibitem{Szegedy2016RethinkingTI}
Christian Szegedy, Vincent Vanhoucke, Sergey Ioffe, Jonathon Shlens, and
  Zbigniew Wojna,
\newblock ``Rethinking the inception architecture for computer vision,''
\newblock {\em 2016 IEEE Conference on Computer Vision and Pattern Recognition
  (CVPR)}, pp. 2818--2826, 2016.

\bibitem{N18-2118}
Chih-Wen Goo, Guang Gao, Yun-Kai Hsu, Chih-Li Huo, Tsung-Chieh Chen, Keng-Wei
  Hsu, and Yun-Nung Chen,
\newblock ``Slot-gated modeling for joint slot filling and intent prediction,''
\newblock in {\em NAACL-HLT}. 2018, pp. 753--757, Association for Computational
  Linguistics.

\bibitem{Pennington2014}
Jeffrey Pennington, Richard Socher, and Christopher~D. Manning,
\newblock ``Glove: Global vectors for word representation,''
\newblock in {\em EMNLP}, 2014.

\bibitem{multijoint}
Dilek Hakkani-T{\"{u}}r, Gokhan T{\"{u}}r, Asli Celikyilmaz, Yun-Nung~Vivian
  Chen, Jianfeng Gao, Li~Deng, and Ye-Yi Wang,
\newblock ``Multi-domain joint semantic frame parsing using bi-directional
  rnn-lstm,''
\newblock June 2016, ISCA.

\bibitem{conneau2017word}
Alexis Conneau, Guillaume Lample, Marc'Aurelio Ranzato, Ludovic Denoyer, and
  Herv{\'e} J{\'e}gou,
\newblock ``Word translation without parallel data,''
\newblock {\em arXiv preprint arXiv:1710.04087}, 2017.

\end{thebibliography}

\end{document}